\pdfoutput=1
\documentclass{article}

\PassOptionsToPackage{numbers,sort&compress}{natbib}
\usepackage[dblblindworkshop, final]{neurips_2026}  
\workshoptitle{Physical World AI: Geometry, Characteristics, and Multimodal Sensing}

\usepackage[utf8]{inputenc} 
\usepackage[T1]{fontenc}    
\usepackage{hyperref}       
\usepackage{url}            
\usepackage{booktabs}       
\usepackage{amsfonts}       
\usepackage{amsmath}
\usepackage{nicefrac}       
\usepackage{microtype}      
\usepackage{xcolor}         
\usepackage{graphicx}

\title{Walks2Maps: Learning Indoor Geometry from Crowdsourced Wi-Fi RTT and Motion Trajectories}

\author{%
  Enguang Fan \quad Binh Minh Tran \quad Fulin Wang \quad Jiaxi Li \quad Klara Nahrstedt \\
  University of Illinois Urbana-Champaign \\
  \texttt{\{enguang2, bmtran2, fulinw2, jiaxili3, klara\}@illinois.edu}
}

\begin{document}

\maketitle

\begin{abstract}
  Everyday smartphone walks offer a scalable source of indoor geometry, but their motion traces drift and their radio measurements are distorted by multipath. We present Walks2Maps, a physics-guided framework that reconstructs a shared metric map from Wi-Fi round-trip-time (RTT) and motion observations, without a floor plan, surveyed access-point (AP) coordinates, or position labels. Walks2Maps jointly estimates pedestrian trajectories, AP locations, and per-trajectory stride scales, then recovers the pathway layout from the reconstructed paths. Shared AP observations connect walks recorded in different local frames, while metric ranges constrain their relative geometry and scale. An asymmetric range loss accommodates multipath-induced excess delay. On 22 phone trajectories collected along 11 routes on an office floor with 30 APs, Walks2Maps achieves median AP and trajectory errors of 0.87\,m and 0.89\,m, respectively, and a walked-area F1 score of 0.985. Ablations show that radio similarity can recognize places while still producing a distorted map; retaining the metric information in RTT resolves this ambiguity. Asymmetric modeling further reduces AP error by 54\% relative to a symmetric loss. These results demonstrate how radio and motion measurements can jointly reveal indoor geometry without a dedicated mapping survey. We will release the dataset and toolkit.
\end{abstract}

\begin{figure}[h]
  \centering
  \includegraphics[width=\linewidth]{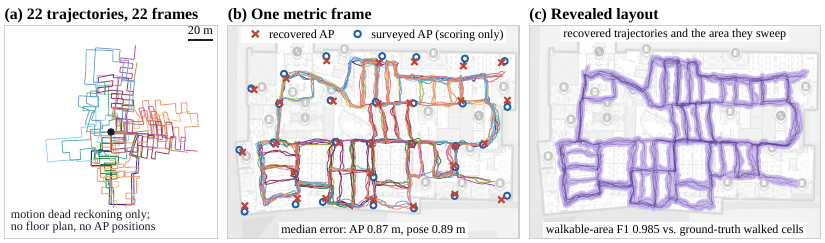}
  \caption{From independent walks to a shared metric map. (a)~Dead-reckoned trajectories of 22 phone recordings, each in its own frame. (b)~Joint estimate; surveyed APs are for evaluation only. (c)~The area within 0.6\,m of the recovered paths reveals the corridor and aisle layout; the floor plan is a visual reference only.}
  \label{fig:teaser}
\end{figure}

\section{Introduction}

Indoor maps support localization, navigation, and spatially aware services, but producing them remains a building-by-building effort: fingerprinting requires a radio survey, range-based positioning requires anchor coordinates, and map matching requires a floor plan. Crowdsourcing offers a scalable alternative by combining trajectories from ordinary walks~\cite{rai2012zee,wang2012unloc,alzantot2012crowdinside,shen2013walkie,gao2014jigsaw}. In many Wi-Fi systems, however, received signal strength (RSS) mainly supplies place associations, such as fingerprint revisits~\cite{rai2012zee,ferris2007wifislam} or Walkie-Markie's WiFi-Marks~\cite{shen2013walkie}, so motion must supply much of the metric structure.

Wi-Fi RTT changes the role of radio. Through IEEE 802.11mc fine timing measurement~\cite{ftm_survey,ibrahim2018verification}, commodity phones measure distances to nearby APs about twice per second in our recordings, so each walk provides a motion trace and metric observations of fixed infrastructure. The challenge is to combine walks in unrelated frames despite uncertain stride lengths and multipath-biased ranges.

We introduce Walks2Maps, a physics-guided framework for reconstructing metric indoor maps from crowdsourced radio and motion traces. It jointly estimates pedestrian trajectories, unknown AP locations, and stride scales, then extracts the pathway layout from the recovered paths (Figure~\ref{fig:teaser}). Our central finding is that accurate place recognition alone does not ensure accurate geometry: metric range constraints establish distances and scale, while propagation-aware loss modeling improves reconstruction under multipath. On an office floor with 30 APs, Walks2Maps achieves median AP and trajectory errors of 0.87\,m and 0.89\,m and recovers 91\% of the reference corridor length.

\section{Method}
\label{sec:method}

\paragraph{Geometric supervision from radio and motion.} Walks2Maps fits a shared geometric model directly to sensor observations: dead reckoning constrains successive positions within each walk, repeated observations connect walks through shared APs, and metric RTT constrains their distances to those APs. No floor plan, surveyed AP coordinate, or ground-truth position is used.

\paragraph{Observations and unknowns.} Trajectory $w$ contains RTT bursts $i=1,\dots,N_w$, each with ranges $r_{wik}$ to APs $k$ and a quality weight $g_{wik}$ (fraction of successful timing frames). Dead reckoning provides the displacement $\Delta q_{wi}$ between consecutive bursts, with unit stride in the trajectory's own frame. We estimate 2-D positions $p_{wi}$ per burst and $a_k$ per AP and, per trajectory, a heading offset $\theta_w$, whose rotation $R(\theta_w)$ maps local odometry into the shared frame, and a positive stride scale $s_w$.

\paragraph{Objective.} We minimize
\begin{equation}
  J=\sum_{w,i}\frac{\lVert p_{w,i+1}-p_{wi}-s_w R(\theta_w)\,\Delta q_{wi}\rVert^2}{\sigma_o^2}
  +\sum_{w,i,k}\rho\!\left(\frac{c(e_{wik})^2\,g_{wik}^2\,e_{wik}^2}{\sigma_r^2}\right)+\Pi(\theta,s).
  \label{eq:J}
\end{equation}
Here $e_{wik}=r_{wik}-d(p_{wi},a_k)$, with $d$ the slant distance under a nominal 1.9\,m AP-to-phone height difference; $\sigma_o$ and $\sigma_r$ weight motion and range residuals, $\rho$ is a soft-$L_1$ robustifier, and $\Pi$ holds weak Gaussian priors on $\theta_w$ and $\log s_w$. We set $c(e)^2=1-\tau$ for $e<0$ and $\tau$ otherwise, with $\tau=0.05$, so before robustification a modeled distance exceeding an observed range receives 19 times the weight of an equally sized positive residual. Inspired by low-expectile regression~\cite{newey1987expectile}, this weighting favors the lower portion of the range observations while accommodating large positive multipath errors. Metric ranges constrain the map's scale and let stride scales be estimated alongside geometry.

\paragraph{Initialization and cross-trajectory association.} For each trajectory-AP pair, the minimum-range burst is a closest-approach candidate; candidates of the same AP seed a rigid registration across trajectories, extending Walkie-Markie's landmarks~\cite{shen2013walkie} with metric ranges. We then refine all trajectories and AP positions jointly by sparse nonlinear least squares with an analytic Jacobian, in about 7\,s overall. For an association-only ablation, we also match five-burst RTT signatures by their median per-AP range discrepancy and group matches into virtual landmarks by leader clustering~\cite{ferris2007wifislam,wei2022rwifislam}, with the matching threshold calibrated on co-carried phones, without position labels.

\paragraph{Pathway reconstruction.} We rasterize the recovered paths on a 0.5\,m grid to estimate the walked area. Cells traversed by at least two trajectories are thinned to a one-cell-wide skeleton, with junctions at cells of three or more neighbors and corridor segments between them; all parameters are fixed before scoring. The map covers the pathways the walks sampled; unvisited space remains unspecified.

\section{Results}
\label{sec:results}

\paragraph{Dataset and evaluation.} We collected data on one office floor of about $75\times40$\,m (Building~A) with 30 RTT-capable APs. Eleven routes, 2.12\,km of walked path, were recorded over four days, each by a person carrying two phones of the same model; the 22 phone trajectories contain 10{,}525 RTT bursts and 100{,}704 ranges at about 1.75 bursts per second. Ground-truth waypoints and surveyed AP coordinates are reserved for evaluation, which uses one rigid alignment and never rescales: joint over the trajectory set, or AP-based for the AP map and range-model ablations (Table~\ref{tab:ablation}).

\begin{figure}[t]
  \centering
  \includegraphics[width=\linewidth]{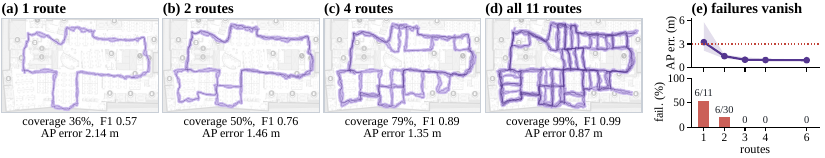}
  \caption{Additional routes improve coverage and reliability. (a--d)~Recovered paths and walked area from the first 1, 2, 4, and all 11 routes (two phones each, solved from scratch); this single route happens to succeed. (e)~Median AP error (interquartile band) and fraction of reconstructions above 3\,m, over all eleven single routes and 30 sampled subsets each of two, three, four, and six routes.}
  \label{fig:crowd}
\end{figure}

\paragraph{Sub-meter geometry from overlapping walks.} Under AP-based alignment, Walks2Maps localizes APs with median (90th-percentile) error 0.87 (1.75)\,m and trajectory positions with 0.89 (1.59)\,m. The recovered walked area achieves F1 0.985 (precision 0.979, recall 0.992; Figure~\ref{fig:teaser}), and its hallway skeleton recovers 91\% of the ground-truth corridor length (535 of 587\,m), with junction precision and recall of 0.73 and 0.67 within 2\,m; most errors arise where nearby parallel aisles merge. A causal particle filter given the surveyed AP coordinates reaches 0.86\,m on the same recordings.

\paragraph{More walks improve coverage and reliability.} The map expands as routes are added (Figure~\ref{fig:crowd}): the illustrated route covers 36\% of the walked area, four routes 79\%, and all eleven 99\%. Median AP error falls from 3.24\,m (single routes) to 0.95\,m (six routes); six of the eleven individual routes fail (median AP error above 3\,m), but none of the 90 tested subsets with three, four, or six routes does.

\begin{table}[t]
  \caption{Contribution of radio information and range modeling (same 22 trajectories and odometry; errors in m). Top: joint rigid alignment of the trajectory set. Bottom: AP-based alignment, which accounts for the two Walks2Maps trajectory errors, 0.83 and 0.89\,m. The particle filter is supplied with surveyed AP coordinates.}
  \label{tab:ablation}
  \centering
  \small
  \begin{tabular}{@{}llccc@{}}
    \toprule
    Merging signal & Method & Pose med. & Pose p90 & Walk.\ F1 \\
    \midrule
    none & odometry + landmark registration & 2.83 & 6.30 & 0.816 \\
    RF similarity & rWiFiSLAM-style closures~\cite{wei2022rwifislam} & 18.28 & 28.90 & 0.028 \\
    RSS landmarks & WiFi-Mark trend reversals~\cite{shen2013walkie} & 11.58 & 17.23 & 0.399 \\
    RTT signatures & virtual landmarks, identity only & 9.68 & 14.98 & 0.388 \\
    \quad + range hinges & virtual landmarks + one-sided hinges & 1.22 & 2.40 & 0.893 \\
    metric RTT ranges & \textbf{Walks2Maps, one-sided (ours)} & \textbf{0.83} & \textbf{1.54} & \textbf{0.990} \\
    \midrule
    & & AP med. & Pose med. & Walk.\ F1 \\
    \midrule
    metric RTT ranges & \textbf{Walks2Maps, one-sided (ours)} & \textbf{0.87} & \textbf{0.89} & \textbf{0.985} \\
    & \quad symmetric loss & 1.89 & 1.36 & 0.871 \\
    & \quad + free per-AP range offset & 2.06 & 2.25 & 0.758 \\
    surveyed APs & causal particle filter (reference) & given & 0.86 & -- \\
    \bottomrule
  \end{tabular}
\end{table}

\begin{figure}[t]
  \centering
  \includegraphics[width=\linewidth]{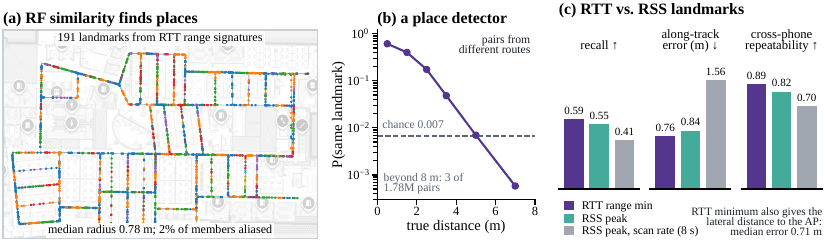}
  \caption{RTT supports both place association and metric landmarks. (a)~Five-burst RTT signatures form 191 virtual landmarks, shown at ground-truth positions for visualization only. (b)~Windows from different routes share a landmark with probability 0.61 within 1\,m, falling to chance (0.007) by 5\,m. (c)~Landmark detection by RTT minima and by RSS peaks (1.6\,Hz and 8\,s scan interval).}
  \label{fig:landmarks}
\end{figure}

\paragraph{Place recognition and metric reconstruction.} Under joint trajectory alignment (Table~\ref{tab:ablation}), odometry with landmark registration alone reaches median error 2.83\,m: each trajectory keeps its dead-reckoned shape at a nominal 0.6\,m stride and is only placed rigidly on shared landmarks. The similarity variants start from such a registration but then optimize poses and strides under place-equality constraints, which do not constrain scale. Our implementations of rWiFiSLAM-style single-burst similarity closures~\cite{wei2022rwifislam} and WiFi-Mark-style RSS landmarks~\cite{shen2013walkie} reach 18.28 and 11.58\,m, the former with loose associations (median true separation 2.47\,m). Five-burst RTT signatures produce more consistent place associations: their 191 virtual landmarks have median radius 0.78\,m, co-carried-phone agreement 0.80 on held-out routes, and typically span seven trajectories from four routes (Figure~\ref{fig:landmarks}a,b). Yet using these landmarks only as place identities yields 9.68\,m, with stride scales contracting to 0.33\,m: with stride left to be estimated, place associations alone do not determine metric distances. One-sided range-difference hinges add a weak metric constraint: by the triangle inequality, two windows must lie at least as far apart as their median range difference to shared APs, less a 0.94\,m slack calibrated on co-carried phones. These lower bounds reduce error to 1.22\,m, and fitting metric RTT ranges directly reaches 0.83\,m and walked-area F1 0.990. Within the range model, a symmetric loss raises AP error from 0.87 to 1.89\,m, and a free per-AP offset to 2.06\,m.

\paragraph{RTT provides repeatable landmarks and distance information.} Across 1{,}626 ground-truth closest-approach events, RTT minima achieve recall 0.59, median along-track error 0.76\,m, and cross-phone repeatability 0.89, against 0.41, 1.56\,m, and 0.70 for RSS peaks at an 8\,s scan interval (Figure~\ref{fig:landmarks}c). RSS at the RTT rate of 1.6\,Hz narrows the gap to 0.55, 0.84\,m, and 0.82, though its precision stays lower (0.58 versus 0.72). Beyond detection, an RTT minimum estimates the closest-approach distance to the AP (median error 0.71\,m), which lets radio constrain the geometry.

\section{Discussion}
\label{sec:discussion}

\paragraph{Relation to prior work.} Prior work has demonstrated crowdsourced pathway mapping~\cite{rai2012zee,wang2012unloc,alzantot2012crowdinside,shen2013walkie,gao2014jigsaw}, RTT-based loop closure~\cite{wei2022rwifislam}, anchor bootstrapping from non-cooperative Wi-Fi ranges~\cite{sie2025crowd}, and meter-level RTT localization on a production WLAN~\cite{fan2026enterprise}. We study the joint reconstruction of pedestrian paths, AP locations, and pathway structure from independently referenced trajectories, with stride scales estimated from the data. Building on asymmetric ranging models~\cite{kok2015uwb,jiang2024robust}, our evaluation isolates how metric distance constraints and propagation-aware loss design contribute to recovering a shared indoor map.

\paragraph{Implications for physical-world learning.} The results show the complementary roles of association and physical measurement: RTT signatures reliably link nearby places, but those links alone leave metric scale unresolved when stride lengths are unknown. These findings suggest a design principle for physical-world learning: place association and metric structure should be modeled together. Similarity identifies recurring observations, while physical measurement constraints establish distances and scale. Propagation-aware losses provide additional geometric supervision, enabling reconstruction directly from unlabeled radio and motion streams.

\paragraph{Limitations and outlook.} The evaluation covers one floor, one AP vendor, and paired recordings from one phone model; the solver runs offline in two dimensions; and the map covers only the corridors and aisles walked. Future work will test heterogeneous devices and buildings, add learned layout priors, span multiple floors, and support online localization against the crowdsourced map.

\begin{ack}
This work used the Delta system at the National Center for Supercomputing Applications [award OAC 2005572] through allocation CIS250086 from the Advanced Cyberinfrastructure Coordination Ecosystem: Services \& Support (ACCESS) program~\cite{boerner2023access}, which is supported by National Science Foundation grants \#2138259, \#2138286, \#2138307, \#2137603, and \#2138296.
\end{ack}

{
\small
\bibliographystyle{unsrtnat}
\bibliography{refs}
}

\end{document}